# A Critical Study towards the Detection of Parkinson's Disease using ML Technologies

Vivek Chetia[1], Abdul Taher Khan[2], Rahish Gogo[3], David Kapsian Khual[4], Dr. Purnendu Bikash[5], Dr. Sajal Saha[6]

Student, The Assam Kaziranga University[1]

Student, The Assam Kaziranga University[2]

Student, The Assam Kaziranga University[3]

Student, The Assam Kaziranga University[4]

Associate Professor, School of Computing Sciences, The Assam Kaziranga University[5]

Professor, Computer Science and Engineering, Adamas University[6]

*vikichet@gmail.com[1], ktaher642@gmail.com[2], gogoirahish@gmail.com[3], kskdev106@gmail.com[4], pbacharyaa@gmail.com[5], sajal1.saha@adamasuniversity.ac.in[6]*

**Abstract**

In this paper, we have implemented a mobile application to detect Early Parkinson's Disease using the Convolutional Neural Network (CNN) model. Using spiral and wave images, with the help of Transfer learning and Image Augmentation, we have trained our model to detect Parkinson's Disease. To generate the result, the mobile application uses the device's camera and storage to access images of spiral and wave drawn by the user. Final result is generated by comparing the result of spiral and wave images. We have achieved accuracy of 95% on Spiral images and 92% on Wave images.

## 1. Introduction

Parkinson's Disease is a disease that occurs with age. It's a disorder in the brain which affects shaking, stiffness, imbalance in walking and coordination. There are many researchers out there who are working on detection of Parkinson's Disease using deep learning. Among all these spiral and wave drawing is one of the uncommon and efficient ways to detect Early Parkinson's Disease using Image Processing. Parkinson's disease is a neurodegenerative disease that causes damage to a core area of the brain due to which patients are not able to draw spiral and wave images properly which is caused by uncontrolled shaking and muscle rigidity that leads to difficulty in movement and performing regular tasks.

In this paper we used a CNN based transfer learning model to classify the early signs of Parkinson's Disease. Here in this project we used MobileNetv2, NasNetMobile, EfficientNetB0, ResNet50, Inceptionv3 models. These models are pre-trained CNN models using ImageNet. These models are low computation cost based models which are specially targeted for mobile hardware.

## 2. Related works

We have gone through various papers related to finding solutions for Parkinson's disease detection with different approaches. We have gone through them and observed our findings that are mentioned in Table 1 including machine learning classification approach to neural networks.

In Paper 1, they have used an audio signal dataset from the UCI dataset repository. A Parkinson's disease patient's noise is low-volume and monotonous. Various audio signals such as jitter, simmer, New Human Revolution (NHR), and Multidimensional Voice Program (MDVP) are used as train and test data. The data is preprocessed using the MinmaxScaler technique. The threshold value and correction coefficient of audio data are used as feature selection parameters.

In Paper 2, they have used speech dataset. KNN, SVM and ELM algorithms were used to train machine learning models to classify patients affected with Parkinson's disease.

In Paper 3, they have used an artificial neural network system using a back propagation method to aid clinicians in detecting Parkinson's disease.

In Paper 4, they have investigated several neural network topologies for detecting Parkinson's Disease using Magnetic Resonance T1 brain images. They proposed three ensemble designs that combine several of the ImageNet Large Scale Visual Recognition Challenge's winning Convolutional Neural Network models.

In Paper 5, keystroke timing data from 103 patients was collected as they wrote on a computer keyboard over a prolonged length of time, demonstrating that Parkinson's Disease affects numerous features of hand and finger movement. To categorise the participants' disease condition, a novel methodology was applied, which included a combination of various keystroke features that were analysed by an ensemble of machine learning classification models.

In Paper 6, they have used various approaches including Classification and Regression, Artificial Neural Networks, and Support Vector Algorithms for the classification of Parkinson's disease.

In Paper 7, they have employed deep learning to categorise motion data recorded in unscripted scenarios from a single wrist-worn IMU sensor. Patients were accompanied by a movement disorder expert for validation purposes, and their motor status was passively monitored every minute.

Table 1. Related Works

| Papers | Findings |
| --- | --- |
| 1 | In this paper, they found that XGBoost accuracy was 96% and Mathews Correlation Coefficient was 89%. |
| 2 | In this paper, KNN, SVM and ELM are used. In which their prediction result shows a difference of 85-95% which shows that speech data can be used for detecting it. |
| 3 | In this paper, they used the ANN model which shows 100% accuracy. |
| 4 | In this paper, they used 6 different models: ResNet 101 accuracy was 92%, SqueezeNet 1.1 accuracy was 60%, DenseNet accuracy was 80%, VGG 19 accuracy was 73%, MobileNet accuracy was 83% and ShuffleNet V2 accuracy was 75%. |
| 5 | 8 models are used for detecting Parkinson's disease with the classification accuracy of 100% |
| 6 | In this paper, they used CART which provided accuray around 85%, SVM accuracy was 79% and ANN accuracy was 91%. |
| 7 | In this paper, KNN, Random Forest, MLP are used with CNN having the highest accuracy of 98% in detecting. |

## 3. Methodology

### 3.1. Data Pre-processing

Dataset contains Spiral and Wave image directory. Each having a 'healthy' and 'parkinson' directory to distinguish between healthy and disease images. Total of 204 images are available, 'healthy' having 102 images and 'parkinson' having 102 images for each Spiral and Wave directory. Since the total images are very less for model training, we have used Image Augmentation which increases the image count by rotating, zoom-in and zoom-out, vertical - horizontal flip. Image Augmentation generated 2000 images, 1000 images each for Spiral and Wave.

Below Figure shows how the data augmentation techniques work with our images.

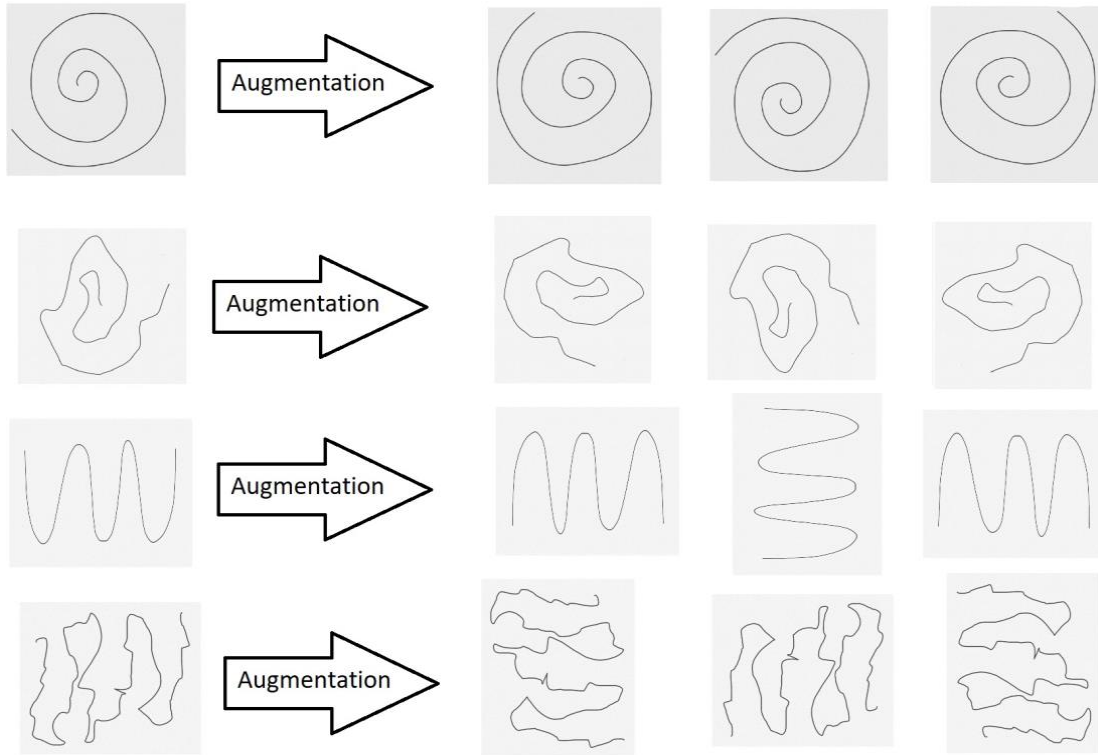

**Figure 1. Data Augmentation**

### 3.2. Model Training and Validation

In this project instead of creating a model from scratch we used a pretrained model using Transfer Learning. This technique is trained specifically to determine a particular classifier in an image. So in this project we implement 5 different models MobileNetV2, NasNetmobile, EfficientNetB0, ResNet50, InceptionV3 which have less computation and light weight as our target is to deploy the model in mobile application. If the model is too heavy the mobile application would not be able to run properly and if the computation is too high it will not be user friendly. Here the Convolutional layer is not trained by the transfer learning; instead, it helps in extracting features for the classification layers.

80% of the dataset is used for model training and 20% of the dataset is used for validating the trained models. Validation Accuracy and Validation Loss is evaluated to determine the best possible model.

### 3.3. Model Testing

After completion of the training of the model we used the test data to determine the accuracy of the model from the result it gives. Here we got an average of 93% accuracy. Once the desired result is obtained the model is converted into tflite format which is used for implementing trained models in mobile applications.

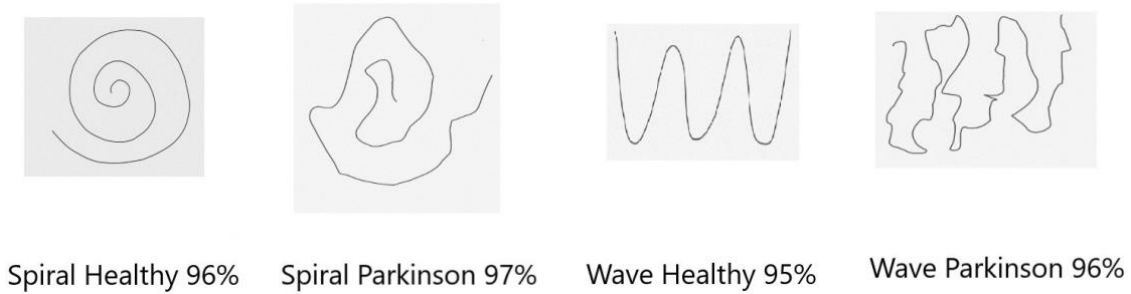

Figure 2. Model Testing Output

### 3.4. Mobile Application Implementation

Using Flutter SDK mobile application is developed so that patients can test for early signs of Parkinson's disease. We used Tensorflow Lite API to access the model in our mobile application.

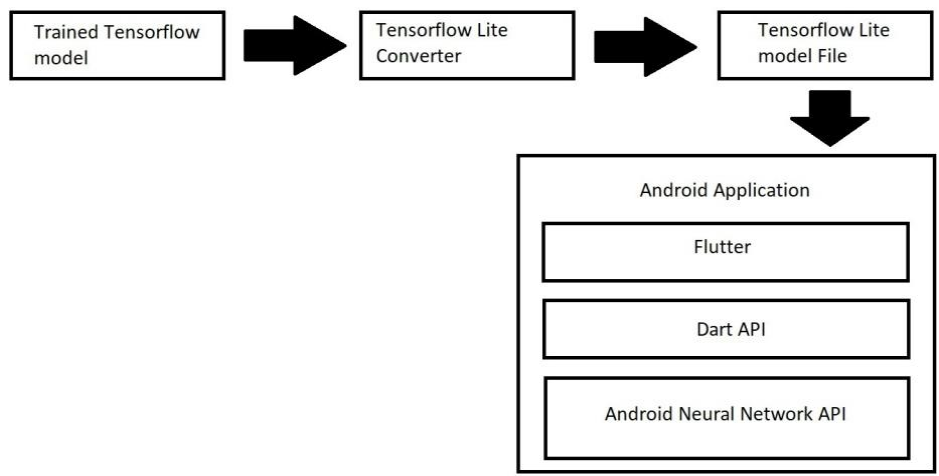

Figure 3. Tensorflow Lite Process

Mobile Application access the device camera or storage to get images of spiral and wave images drawn by the user. Model is accessed using the Tensorflow lite API to get the result. Model gives back the result for both spiral and wave images. Label is then decided by comparing the percentage value for 'healthy' and 'parkinson'. Highest value will be taken as a final result. Final result will be displayed by the application including the percentage value along with the label that is 'Healthy' or 'Parkinson'.

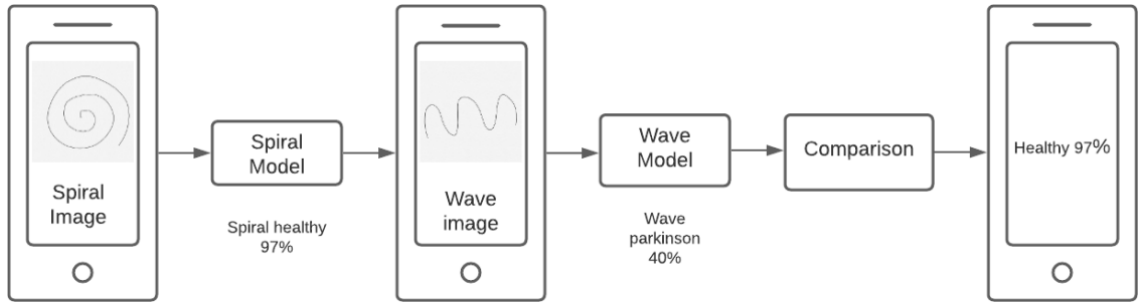

Figure 4. Mobile Application implementation

## 4. Comparison And Result

From the Fig. 5 below we can see that in our experiment on the model trained on spiral images the most accuracy and less loss came up to be MobileNetV2. With a validation accuracy of 0.95 and validation loss of 0.11. Next followed by InceptionV3 with 0.93 accuracy and 0.27 loss. Here RestNet50 performed the worst with validation accuracy of 0.73 and validation loss of 0.53.

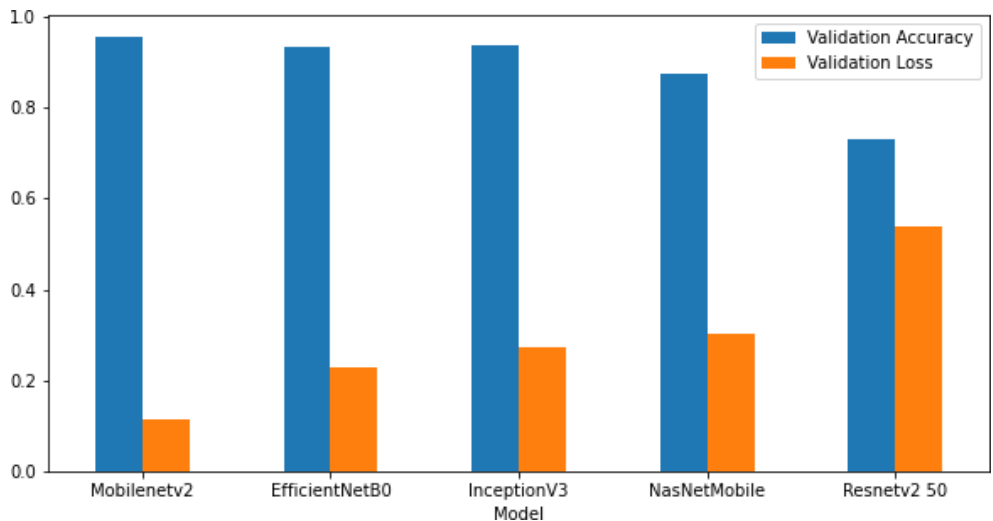

Figure 5. Spiral Model Result

Fig. 6 shows the result of an experiment done on the model trained on wave images the most accuracy and less loss came up to be MobileNetV2. With validation accuracy 0.92 and validation loss 0.18. Next best performing is the EfficientNetB0 model with validation accuracy of 0.91 and validation of 0.15. Here too we can see the ResNet50 did not perform well coming up with validation accuracy of 0.83 and validation loss of 0.42.

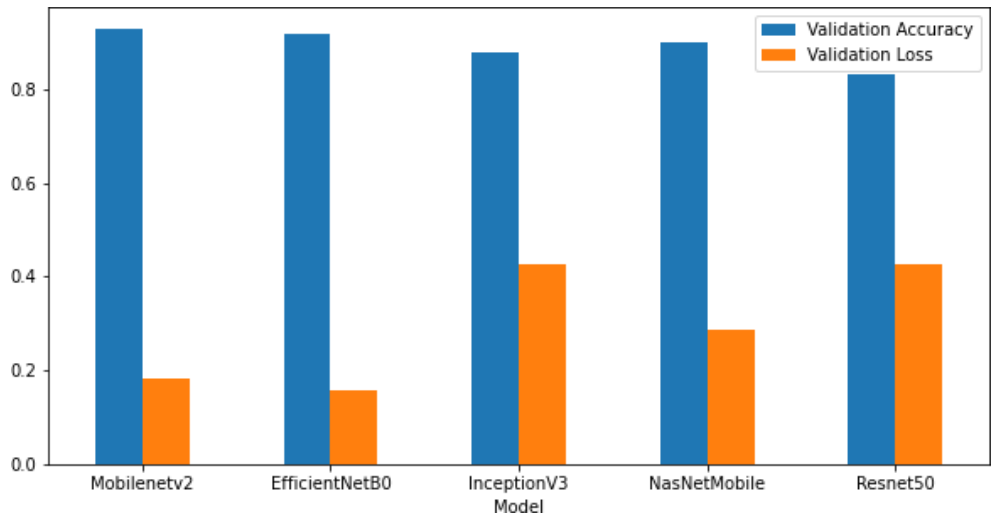

Figure 6. Wave Model Result

Each model is trained for 100 epochs. Fig. 7 illustrates the learning curve of the training accuracy for the model trained on spiral images. Fig. 8 illustrates the learning curves of the validation accuracy for the model trained on spiral images. The models were evaluated on the validation data after each epoch. Compared to the training curves, the validation curve is uneven.

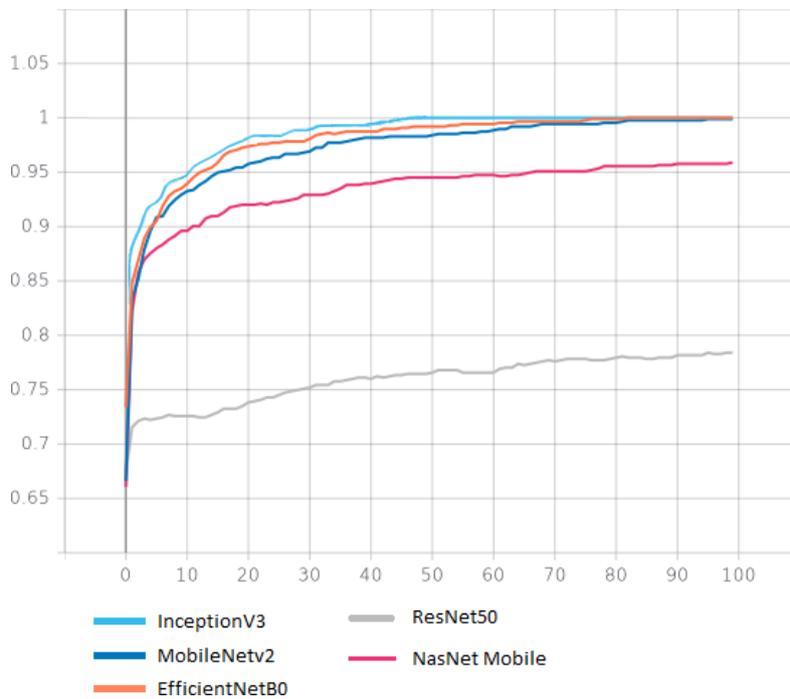

Figure 7. Spiral Model Training Accuracy

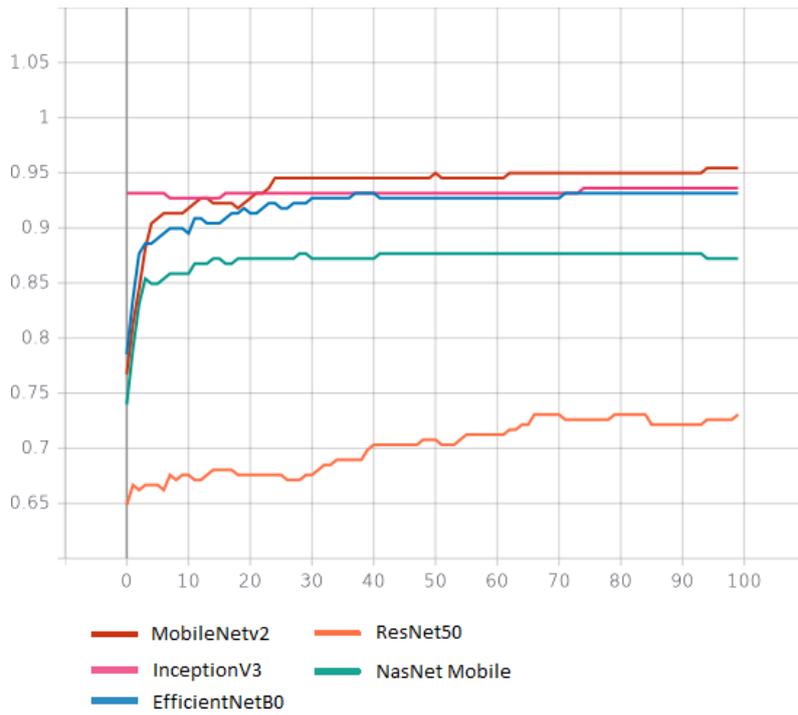

Figure 8. Spiral Model Validation Accuracy

Fig. 9 illustrates the learning curve of the training accuracy for the model trained on wave images. Fig. 10 illustrates the learning curve of the validation accuracy for the model trained on wave images.

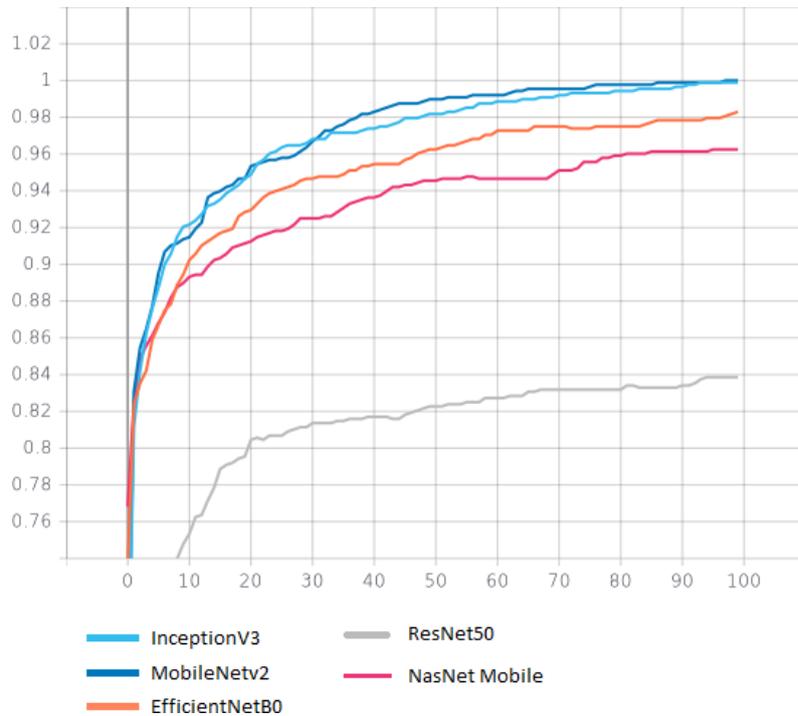

Figure 9. Wave Model Training Accuracy

We can see the similar results for both spiral and wave models. In both cases MobileNetv2 takes the first place followed by InceptionV3 and EfficientNetB0. We can observe from the validation curves of both spiral and wave models in the range between 25-60 epochs there is no improvement and later can be taken as the situation of overfitting. So, in order to avoid overfitting of the model, the Early Stopping function is used to train the model again to stop the training process after the next 3 epochs if there is no improvement in the accuracy.

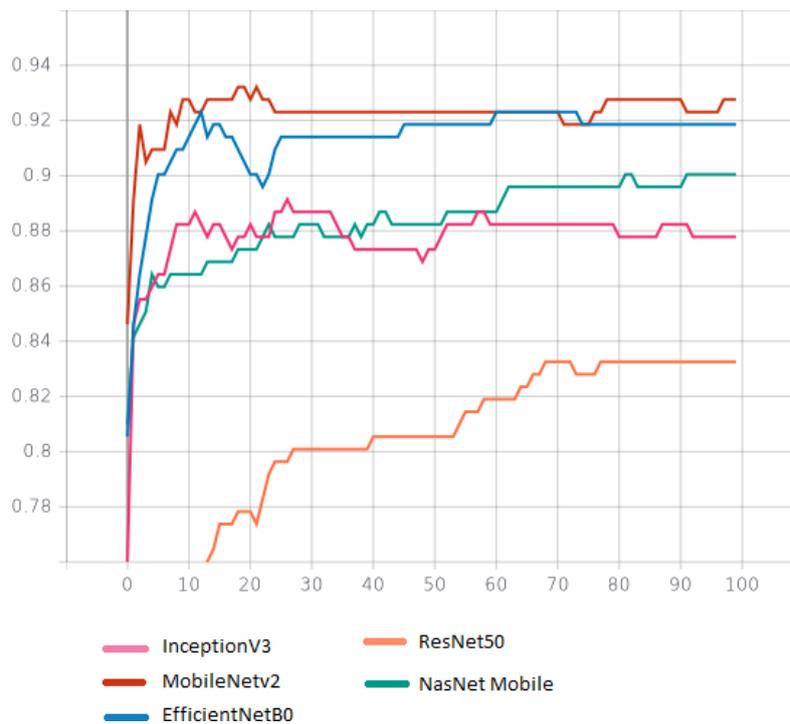

Figure 10. Wave Model Validation Accuracy

## 5. Conclusion

To conclude our work, the mobile application will be installed into the device[8],[9]. It will use the phone camera to capture the drawn spiral and wave image. The images will go through the install model for image classification, once done the output will be shown in the mobile screen.

To create the model, at first, we preprocessed all the datasets for deep learning. As due to less availability of data we go through data augmentation, using the python library "imgaug". And for classification of images, we used a pretrained model called transfer learning which works in Keras with weight "ImageNet". Once the model is built, the model is converted into a mobile deployable file called FlatBuffer (.tflite). This converts TensorFlow Lite into a mobile file to deploy in mobile devices. After the conversion of (.tflite) it is then converted into a .apk file that is an executable file for android devices. The model can be improved by having a more original data set instead of using data augmentation.